\documentclass[letterpaper, 10 pt, conference]{ieeeconf}  

\IEEEoverridecommandlockouts                              

\usepackage{graphics} 
\usepackage{epsfig} 
\usepackage{mathptmx} 
\usepackage{times} 
\usepackage{amssymb}  
\usepackage{amsmath}
\usepackage{amsfonts}
\usepackage{graphicx}
\usepackage{booktabs}
\usepackage{multirow}
\usepackage{wrapfig}
\usepackage{threeparttable}
\usepackage{xcolor}
\usepackage{arydshln}

\title{\LARGE \bf%
Taming an End-to-End Autonomous Driving Policy \\for Urban Navigation of Quadruped Robots
}

\author{
Joochan Kim$^{1*\bullet}$
Chanuk Yang$^{2*}$
Tackgeun You$^{2}$
Ziran Wang$^{1}$
Hwasup Lim$^{2}$%
    \thanks{$^{1}$Purdue University}%
    \thanks{$^{2}$Korea Institute of Science and Technology (KIST)}%
    \thanks{*Authors have equal contributions}%
    \thanks{$^{\bullet}$Work was done when the author was at KIST}
}

\begin{document}

\maketitle
\thispagestyle{empty}
\pagestyle{empty}

\begin{abstract}
We present Go2-DrivoR, a goal-conditioned adaptation of the end-to-end autonomous driving trajectory planning framework DrivoR for urban navigation with quadrupedal robots.
By conditioning trajectory generation on a local-frame subgoal through a goal token and adapting the vehicle-centric scoring formulation, the method extends DrivoR to short-horizon goal-conditioned local planning without redesigning its core decoders.
Specifically, we redefine drivable-area compliance for sidewalk-oriented navigation and reformulate the original ego progress term as goal-conditioned ego progress.
Trained exclusively on TartanGround simulation data, Go2-DrivoR improves waypoint-conditioned planning performance on unseen simulation environments and transfers zero-shot to open-loop real-world trajectory prediction.
\end{abstract}
\section{INTRODUCTION}
\label{sec:intro}
Urban navigation with quadrupedal robots requires short-horizon goal-conditioned local planning under legged motion constraints.
In contrast to autonomous vehicles, quadrupedal robots operate in roadways, walkways, and mixed urban environments where lane centerlines and fixed route geometries are not generally available.
A navigation policy must therefore generate local trajectories from visual observations and proprioceptive states while accounting for obstacles, pedestrians, surface variation, and the tracking limitations of a downstream locomotion system.
These characteristics make the direct application of vehicle-centric planning methods nontrivial~\cite{Lee_2024,zhou2025ainavlargelanguagemodelbased}.

Recent end-to-end autonomous driving methods integrate learned scene representations with trajectory planning~\cite{Chitta2023PAMI, sun2025sparsedrive}.
DrivoR is particularly relevant to this setting because it combines register-token-based visual encoding with a disentangled trajectory-scoring architecture for multi-camera planning~\cite{kirby2026drivor}.
However, DrivoR is designed for vehicle-centric environments, where candidate trajectories are evaluated using road-aligned notions of drivable-area compliance and progress~\cite{Dauner2024NEURIPS,sun2026sparsedrivev2}.
Although its safety-related terms can be retained with robot-footprint-aware evaluation, the original drivable-area compliance criterion does not encode the relative desirability of sidewalks, roads, and out-of-bounds regions for quadrupedal navigation.
Moreover, the original formulation lacks explicit goal conditioning; it neither conditions trajectory proposals on a local lookahead waypoint nor measures progress toward that target.

To address this problem, we adapt DrivoR to short-horizon goal-conditioned quadrupedal navigation while retaining its core decoder architectures.
We condition trajectory proposals on a local-frame lookahead waypoint through a goal token.
For trajectory scoring, we retain the existing safety criteria with robot-footprint-aware evaluation and redefine drivable-area compliance (DAC) using sidewalk, road, and out-of-bounds regions.
We also augment the original ego-progress (EP) term with goal-directed progress to obtain goal-conditioned ego progress (GEP), aligning trajectory evaluation with the supplied local goal.

\begin{figure*}[t]
    \centering
    \includegraphics[width=0.95\textwidth]{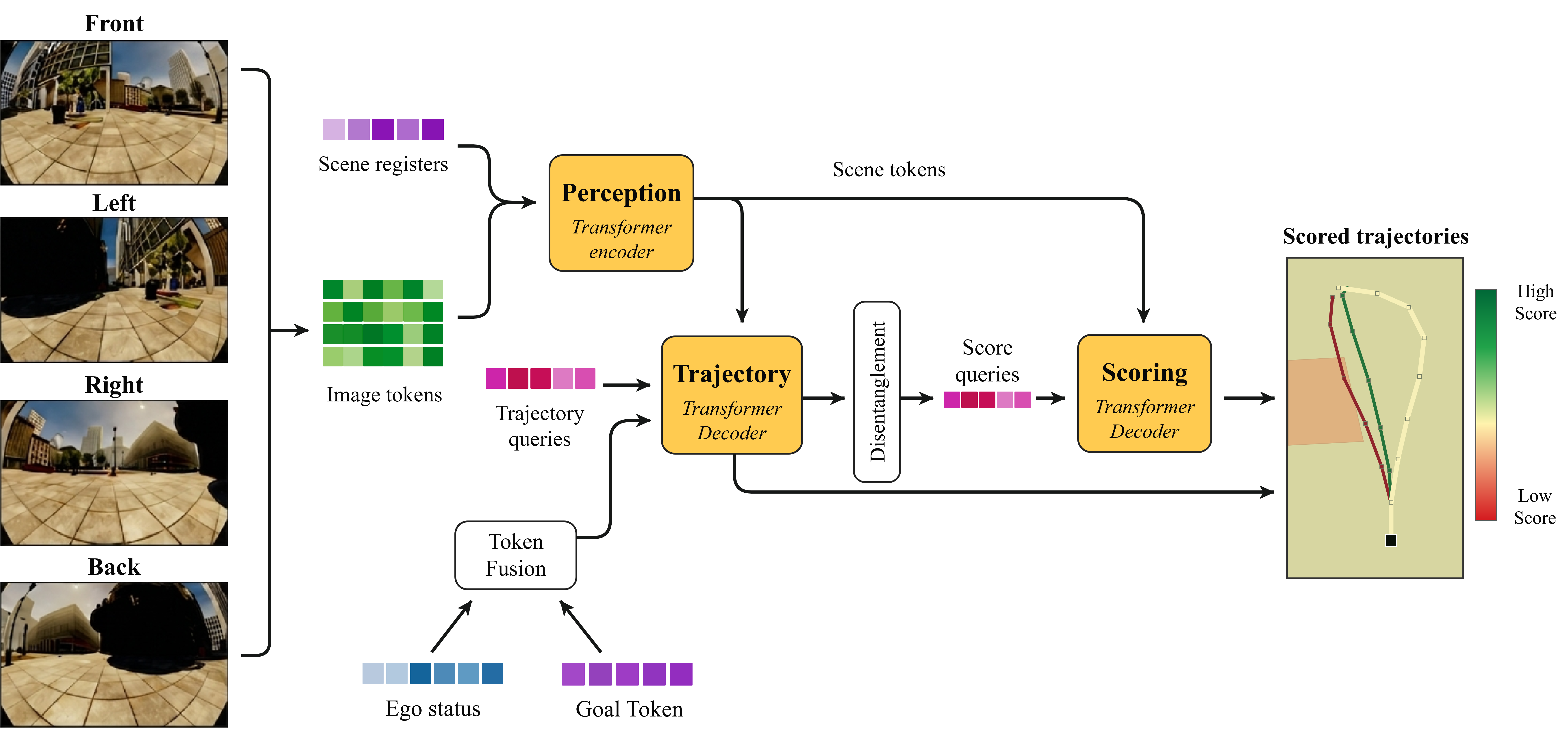 }
    \caption{\textbf{Overall framework of Go2-DrivoR.}
    Multi-camera fisheye images are encoded into scene tokens, while fused ego-state and goal tokens condition trajectory queries.
    The generated trajectories are ranked using adapted scoring criteria, retaining DrivoR's trajectory and scoring decoder architectures.
    }
    \label{fig:overview}
\end{figure*}

To assess sim-to-real transfer, we train the model exclusively on simulation data from TartanGround~\cite{patel2025tartanground}.
Real-world logs collected with the Unitree Go2 are withheld from training and used solely for zero-shot open-loop trajectory evaluation.
We evaluate the model on seen and unseen simulation environments and real-world Unitree Go2 logs.
This protocol evaluates the transferability of the proposed goal-conditioned adaptation without real-world fine-tuning~\cite{bjelonic2025bridginggapsystematicsimtoreal}.
\section{METHODS}
\label{sec:methods}

\subsection{Go2-DrivoR}
We formulate urban quadrupedal navigation as goal-conditioned local trajectory planning.
Given four surrounding fisheye images, the robot's proprioceptive state, a high-level directional command, and a two-dimensional goal in the current ego frame,
Go2-DrivoR generates $K$ candidate trajectories and selects the one with the highest predicted score.
The output consists of eight $(x,y,\theta)$ waypoints over a four-second horizon, expressed in the current ego frame.

Our model builds on DrivoR~\cite{kirby2026drivor}, retaining its visual feature extractor, trajectory decoder, scoring decoder, and winner-takes-all trajectory objective.
We introduce goal-token conditioning for trajectory generation and adapt the scoring criteria through sidewalk-oriented drivable-area compliance and goal-conditioned ego progress (GEP).

\subsubsection{Ego-Proprioceptive and Goal Token Extraction}
We gather the robot's physical status $\mathbf{s}_{\text{ego}} \in \mathbb{R}^{S}$ containing proprioceptive signals, including relative ego pose, linear and angular velocities, acceleration, and command inputs.
This state is projected by a linear layer to obtain the ego status token $\mathbf{e}_{\text{ego}} \in \mathbb{R}^{D}$.
For each local planning sample, we compute a two-dimensional goal $\mathbf{g}=(x,y)\in\mathbb{R}^{2}$ by transforming the waypoint four seconds after the current timestamp into the current ego-centric frame.
The coordinates are normalized by a distance scale
$d_{\text{scale}}=4\,\mathrm{m}$:
\begin{equation}
\bar{\mathbf{g}} =
\left(
\frac{x}{d_{\text{scale}}},
\frac{y}{d_{\text{scale}}}
\right)
\in \mathbb{R}^{2}.
\end{equation}
The normalized subgoal is projected into the latent space to obtain the goal token:
\begin{equation}
\mathbf{e}_{\text{goal}}
=
\mathbf{W}_{\text{goal}}\bar{\mathbf{g}}
+
\mathbf{b}_{\text{goal}},
\end{equation}
where $\mathbf{W}_{\text{goal}}\in\mathbb{R}^{D\times 2}$ and $\mathbf{b}_{\text{goal}}\in\mathbb{R}^{D}$ are learnable parameters.

\begin{table*}[t]
\centering
\caption{
\textbf{Open-loop evaluation on TartanGround.}
Best values, excluding the GT agent, are bold.
}
\label{tab:ol_sim}
\resizebox{0.9\textwidth}{!}{%
\begin{tabular}{lcclccccccc}
\toprule
Split & \# Maps & \# Scenarios & Method &
& NOC $\uparrow$
& TTC $\uparrow$
& DAC $\uparrow$
& GEP $\uparrow$
& Comf. $\uparrow$
& Score $\uparrow$ \\
\midrule

\multirow{4}{*}{Seen}
& \multirow{4}{*}{6}
& \multirow{4}{*}{2,554}
& GT Agent &
& 0.9636 & 0.9318 & 0.9954 & 1.0000 & 0.9405 & 0.9304 \\
\cdashline{4-11}
&&& Constant Vel. &
& 0.8728
& 0.9277
& 0.9543
& 0.5339
& \textbf{1.0000}
& 0.6953 \\
&&& iPlanner &
& 0.8978
& 0.9046
& 0.9838
& 0.7690
& 0.5932
& 0.7344 \\
&&& Go2-DrivoR &
& \textbf{0.9397}
& \textbf{0.9391}
& \textbf{0.9869}
& \textbf{0.7875}
& 0.9894
& \textbf{0.8368} \\

\midrule

\multirow{4}{*}{Unseen}
& \multirow{4}{*}{3}
& \multirow{4}{*}{2,707}
& GT Agent &
& 0.9605 & 0.9522 & 0.9956 & 0.9950 & 0.9102 & 0.9269 \\
\cdashline{4-11}
&&& Constant Vel. &
& 0.8973
& 0.9495
& 0.9032
& 0.5957
& \textbf{1.0000}
& 0.7425 \\
&&& iPlanner &
& 0.9372
& 0.9400
& 0.9836
& \textbf{0.7827}
& 0.5962
& 0.7721 \\
&&& Go2-DrivoR &
& \textbf{0.9579}
& \textbf{0.9539}
& \textbf{0.9888}
& 0.7757
& 0.9907
& \textbf{0.8512} \\

\bottomrule
\end{tabular}%
}
\end{table*}

\paragraph{Goal Token Fusion and Trajectory Query Conditioning}
The ego status token and goal token are concatenated and projected back to the model dimension:
\begin{equation}
\tilde{\mathbf{e}}_{\text{ego}}
=
\mathbf{W}_{\text{fuse}}
[\mathbf{e}_{\text{ego}};\mathbf{e}_{\text{goal}}]
+
\mathbf{b}_{\text{fuse}}.
\end{equation}
We retain the original winner-takes-all trajectory training objective of DrivoR.
Each learnable trajectory query is conditioned by residual addition of the fused token:
\begin{equation}
\tilde{\mathbf{q}}_k
=
\mathbf{q}_k
+
\tilde{\mathbf{e}}_{\text{ego}},
\quad k\in\{1,\ldots,K\}.
\end{equation}
The conditioned queries are passed to the unchanged trajectory decoder, which generates $K=64$ candidate trajectories $\tau_k \in \mathbb{R}^{n_p \times 3}$.
Following DrivoR, each trajectory comprises $n_p=8$ future $(x,y,\theta)$ waypoints sampled at 0.5-second intervals over a four-second horizon.

\subsubsection{Trajectory Scoring for Urban Navigation}
\label{sec:scores}
Compared with the original six-component scoring formulation of DrivoR, our adapted scorer uses five supervised components: no-fault collision (NOC), drivable-area compliance (DAC), time-to-collision (TTC), goal-conditioned ego progress (GEP), and comfort.
The driving-direction compliance (DDC) component is omitted because quadrupedal robots are not constrained to follow a prescribed road or lane direction during destination-based navigation.
The following paragraphs describe the robot-footprint-aware safety evaluation and the adapted DAC and GEP formulations.

\paragraph{Safety and Drivable-Area Criteria}
We retain DrivoR's NOC and TTC formulations but replace the vehicle footprint with the quadrupedal robot footprint for collision and time-to-collision evaluation.
To reflect sidewalk-oriented navigation, we redefine DAC using semantic regions along each candidate trajectory.
A candidate receives a DAC score of $1$ if all sampled waypoints lie on the sidewalk, $0.5$ if at least one waypoint lies on the road while no waypoint lies outside the allowable region, and $0$ otherwise.
These semantic-region labels are used to construct the DAC oracle targets for scoring supervision.

\begin{figure*}[!t]
\centering
\includegraphics[width=\textwidth]{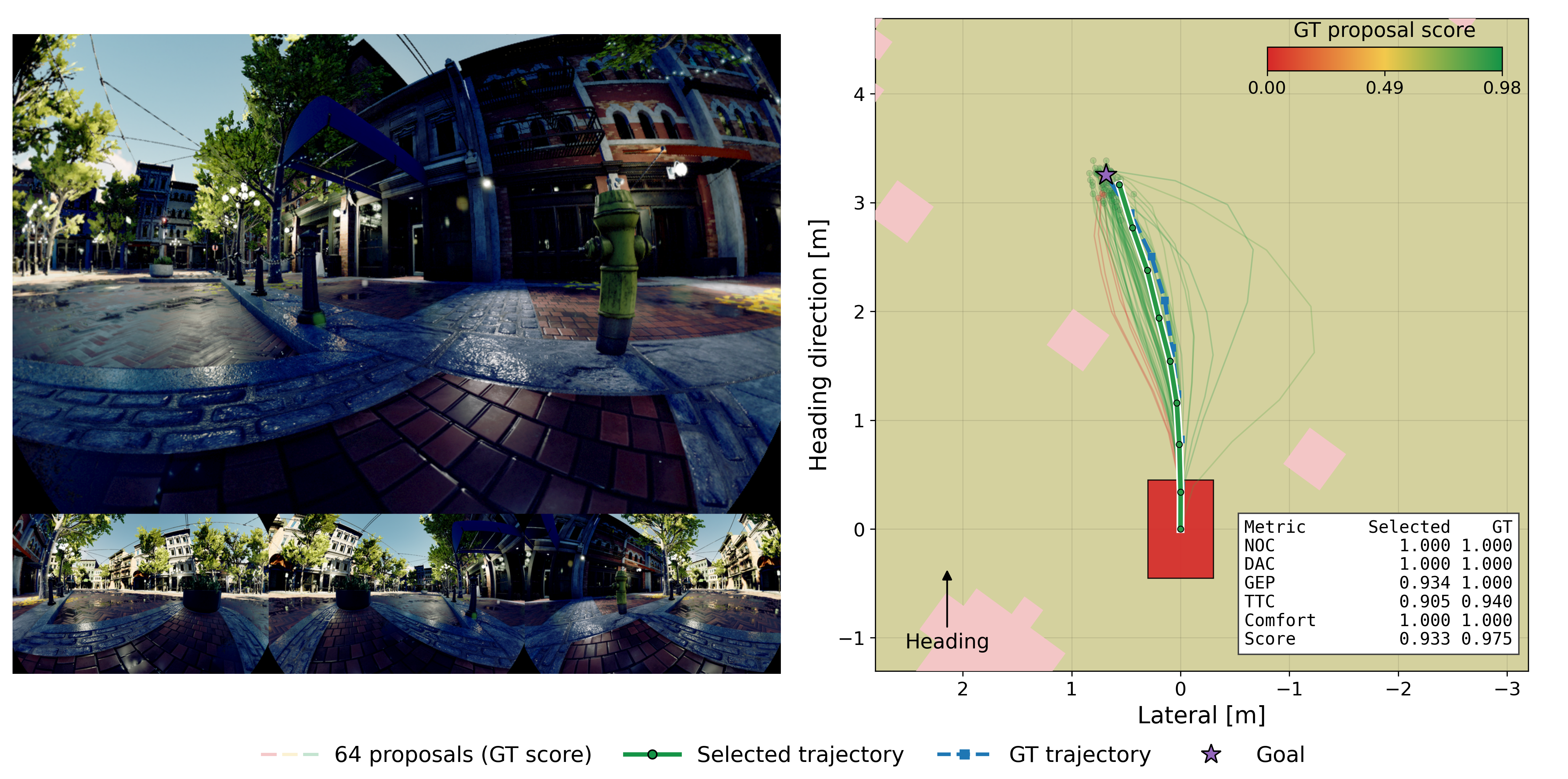}
\caption{\textbf{Open-loop results on TartanGround.}
The left panel shows the input fisheye images: front at the top, and left, rear, and right views from left to right along the bottom.
The right panel shows candidate trajectories colored by oracle-evaluated PDMS-style scores, the selected trajectory in \textcolor{green}{green}, the expert trajectory in \textcolor{blue}{dashed blue}, and the goal marked by a \textcolor{purple}{star}.
\textcolor{yellow}{Yellow} denotes walkable ground, and \textcolor{pink}{pink } denotes non-traversable areas.
The \textcolor{red}{red} rectangle indicates the robot footprint.
The ``Selected'' and ``GT'' columns report oracle-evaluated
sub-scores and aggregate scores for the selected and expert
trajectories, respectively.
For the selected trajectory, the scoring decoder predicts NOC $=0.993$, DAC $=1.000$, TTC $=0.994$, GEP $=0.924$, and comfort $=1.000$, yielding an aggregate PDMS-style score of $0.959$.
}
\label{fig:ol_sim}
\end{figure*}

\paragraph{Goal-Conditioned Ego Progress (GEP)}
To explicitly account for progress toward a specified local
destination, we augment the original ego-progress term with
a goal-directed progress (GP) component.
Given a goal $\mathbf{g}\in\mathbb{R}^{2}$ in the current
ego frame, this component measures the candidate trajectory's
endpoint displacement projected onto the goal direction:
\begin{equation}
    \texttt{goal}(\tau_k) = \text{clip}\left(
    \frac{(\mathbf{p}^{(n_p)}_k - \mathbf{p}^{(0)}_k)\cdot \hat{\mathbf{g}}}
    {\min(\|\mathbf{g}\|_, d_{\text{scale}})},
    0, 1
    \right),
\label{eq:goal}
\end{equation}
where $\hat{\mathbf{g}}=\mathbf{g}/\|\mathbf{g}\|_2$ denotes the unit vector pointing toward the goal, $\mathbf{p}^{(0)}_k=\mathbf{0}$ is the current robot position in the ego frame, and $\mathbf{p}^{(t)}_k\in\mathbb{R}^{2}$, $t=1,\ldots,n_p$, denotes a predicted future position at future time step $t$.

The GEP target is defined by combining the original ego-progress target with the goal-directed progress value:
\begin{equation}
    \texttt{gep}(\tau_k)
    =
    (1-\alpha)\texttt{ep}(\tau_k)
    +
    \alpha\texttt{goal}(\tau_k),
\label{eq:gep}
\end{equation}
where $\alpha \in [0,1]$ and $\alpha=0.5$ in our experiments unless specified.
This formulation retains the original progress signal
while additionally rewarding motion toward the local goal.

\paragraph{Composite Ranking Score}
At inference, the predicted sub-scores are combined using the log-domain formulation of the original DrivoR scorer. Let $\ell_{k,c}$ denote the predicted logit for candidate trajectory $\tau_k$ and score component $c$, and let
\begin{equation}
    \hat{m}_{k,c} = \sigma(\ell_{k,c})
\end{equation}
denote the corresponding sigmoid-transformed sub-score. Since DDC is omitted and GEP replaces the original EP term, the composite score is computed as
\begin{equation}
\begin{aligned}
    s(\tau_k) ={}&
    \log\!\left(\hat{m}_{k,\mathrm{noc}}\right)
    + \log\!\left(\hat{m}_{k,\mathrm{dac}}\right) \\
    &+ \log\!\left(
        5\hat{m}_{k,\mathrm{ttc}}
        + 5\hat{m}_{k,\mathrm{gep}}
        + 2\hat{m}_{k,\mathrm{comfort}}
    \right).
\end{aligned}
\label{eq:comp}
\end{equation}
The composite score is used for proposal ranking and is not directly optimized; the scoring heads are supervised independently using component-wise BCE-with-logits losses.

For evaluation, we compute an adapted PDMS-style aggregate
score from oracle-evaluated sub-scores:
\begin{equation}
\begin{aligned}
A(\tau)
&= \frac{\mathrm{NOC}(\tau)\mathrm{DAC}(\tau)}{12} \\
&\quad\times
\bigl(5\,\mathrm{TTC}(\tau)+5\,\mathrm{GEP}(\tau)
+2\,\mathrm{Comfort}(\tau)\bigr).
\end{aligned}
\label{eq:aggregate}
\end{equation}

\begin{table*}[t]
\centering
\caption{\textbf{Zero-shot performance on real-world dataset.}
}
\resizebox{\textwidth}{!}{%
\begin{tabular}{l c l c c c c c c}
\toprule
Scene & \# Scenarios & Method &
& ADE (m) $\downarrow$ 
& FDE@1s (m) $\downarrow$ 
& FDE@2s (m) $\downarrow$ 
& FDE@4s (m) $\downarrow$ 
& Heading MAE ($^\circ$) $\downarrow$ 
\\
\midrule

\multirow{2}{*}{Pedestrian}
& \multirow{2}{*}{5990}
& Constant Vel. &
& 0.785
& 0.318
& 0.672
& 1.470
& 8.94
 \\
& & Go2-DrivoR &
& \textbf{0.200}
& \textbf{0.149}
& \textbf{0.233}
& \textbf{0.209}
& \textbf{5.79}
\\
\midrule

\multirow{2}{*}{Road}
& \multirow{2}{*}{719}
& Constant Vel. &
& 0.824
& 0.333
& 0.704
& 1.542
& 9.22
\\
& & Go2-DrivoR &
& \textbf{0.213}
& \textbf{0.164}
& \textbf{0.263}
& \textbf{0.186}
& \textbf{6.81}
\\
\bottomrule
\end{tabular}
}
\label{tab:ol_real}
\end{table*}

\paragraph{Scoring Loss}
The scoring decoder is trained using component-wise
binary cross-entropy-with-logits losses:
\begin{equation}
    \mathcal{L}_{\mathrm{score}}
    =
    \sum_{c \in \mathcal{C}}
    \frac{1}{K}
    \sum_{k=1}^{K}
    \mathrm{BCEWithLogits}
    \left(\ell_{k,c}, m_{k,c}\right),
\end{equation}
where $\mathcal{C}$ is the set of supervised components, and $m_{k,c}$ and $\ell_{k,c}$ are the oracle target and predicted logit for candidate $\tau_k$ and component $c$.

We jointly optimize the trajectory and scoring losses with equal weights. Both update the shared perception encoder, but scoring gradients do not pass through predicted trajectories to the trajectory decoder.

\subsection{Simulation and Real-World Data Collection}
\label{subsec:data}
We train the planner on high-fidelity simulation data and evaluate its zero-shot sim-to-real transfer on real-world Unitree Go2 logs.
Both domains undergo a standardized preprocessing and serialization pipeline to unify sensor modalities, coordinate systems, and high-level routing interfaces into the planner's input format.

\begin{figure}[!t]
\centering
\includegraphics[width=\columnwidth]{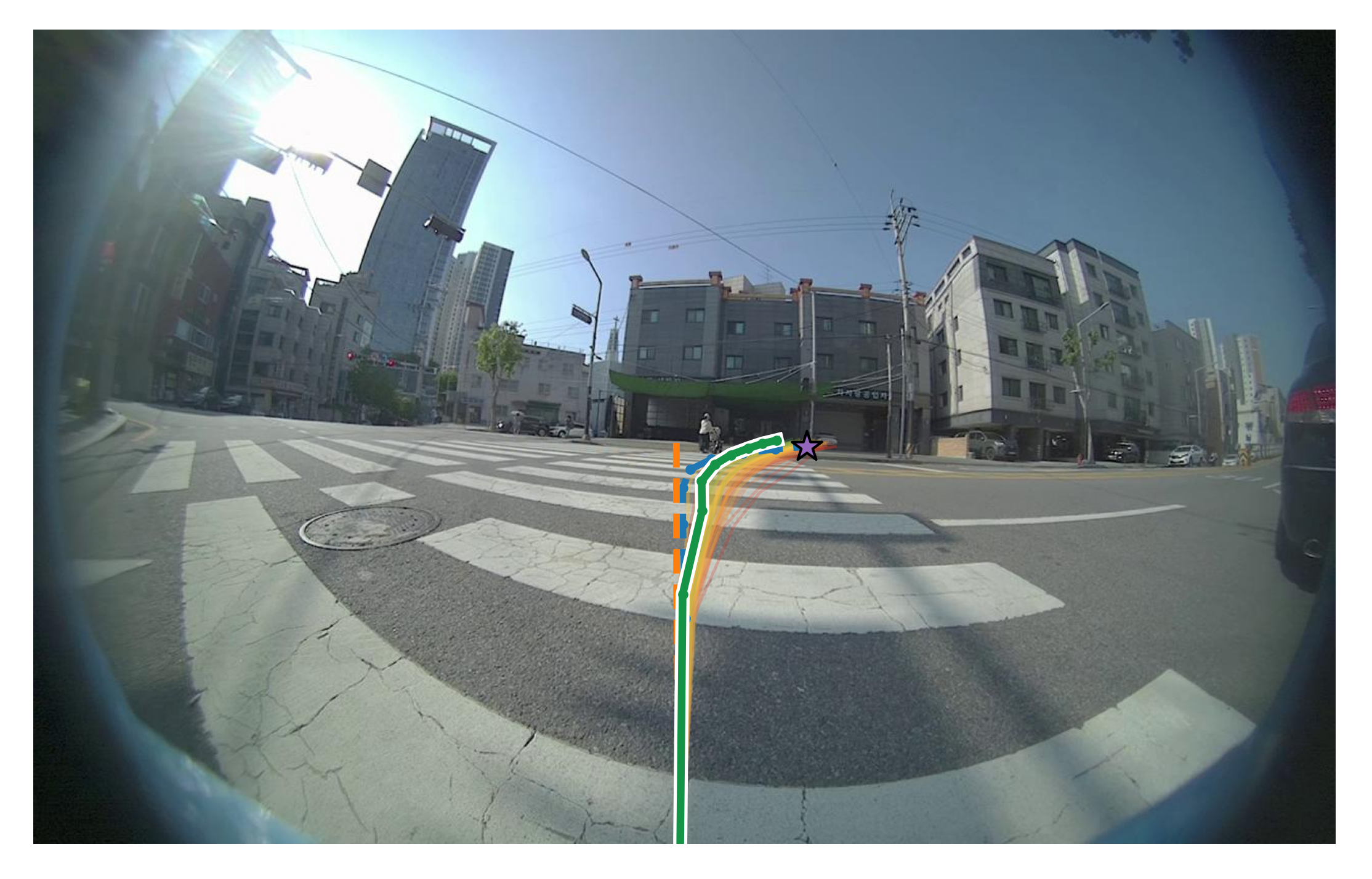}
\caption{\textbf{Qualitative results on real-world open-loop evaluation.}
The expert trajectory, selected trajectory, and constant-velocity prediction are shown in \textcolor{blue}{dashed blue}, \textcolor{green}{green}, and \textcolor{orange}{dashed orange}, respectively.
\textcolor{purple}{Purple stars} mark the four-second lookahead goals.
For the selected trajectory, the scoring decoder predicts NOC $=0.993$, DAC $=0.754$, TTC $=0.916$, GEP $=0.906$, and comfort $=1.0$, with an aggregate PDMS-style score of $0.693$.
}
\label{fig:ol_real}
\vspace{-1.5mm}
\end{figure}

\subsubsection{TartanGround}
TartanGround~\cite{patel2025tartanground} serves as our simulation dataset, providing photorealistic multi-camera renderings, IMU telemetry, and dense semantic point clouds across diverse synthetic environments.
We process the data through a three-stage pipeline.
First, the original multi-camera pinhole images provided by TartanGround are reprojected into four fisheye views---front, left, right, and rear---using a calibrated double-sphere camera model.
Second, the resulting images are paired with expert kinematics, and pose, velocity, and body acceleration are transformed from the simulation coordinate system to ENU (East--North--Up) and FLU (Front--Left--Up) frames.
To follow the NAVSIM convention, we convert the long continuous TartanGround trajectories into local planning samples using a sliding-window procedure.
For a sample at time t, the local-frame goal is defined as the expert pose four seconds after t and is expressed in the current ego frame.
This four-second lookahead waypoint is used as the conditioning goal for the corresponding local planning sample, rather than as a global destination.
High-level four-way commands are then synthesized from the lateral displacement of the same four-second lookahead waypoint in the ego frame.
Finally, semantic point clouds are projected onto a 2D grid to construct boundary, roadway, and walkway vector layers for scoring.

\subsubsection{Real-world}
\label{subsubsec:data_real}
For real-world data collection, the Unitree Go2 is equipped with four GMSL Owl cameras mounted at $90^\circ$ azimuthal intervals around the robot, providing front, left, rear, and right views.
A human operator teleoperated the robot during data collection, and the recorded robot motion serves as the expert trajectory for open-loop evaluation.
The collected logs consist of multi-camera video streams and synchronized state records indexed by frame markers.
We align the visual streams with the proprioceptive logs via temporal cross-referencing, and the raw fisheye images are used directly as input without rectification.
The kinematic states are converted from the robot's native FRD (Front-Right-Down) frame to FLU, and goal-oriented commands are inferred from lookahead targets using a lateral threshold before serialization.
\section{EXPERIMENTS}
\label{sec:exp}

\subsection{Implementation Details} 
The training data consist of 14,192 local planning samples from the TartanGround dataset~\cite{patel2025tartanground}.
We train the model for 20 epochs using AdamW~\cite{loshchilov2018decoupled} on three NVIDIA H200 GPUs with a global batch size of 48.
The learning rate is set to $5\times10^{-4}$, with a two-epoch warm-up phase, followed by cosine annealing.
Open-loop inference is performed on a single NVIDIA H200 GPU, with measured batch latencies of 10.8\,ms at batch size 1 and 94.1\,ms at batch size 16.

\begin{table*}[!t]
    \centering
    \caption{\textbf{Ablation study on goal-token conditioning and progress blending.}
    }
    \label{tab:goal_ablation}
    \resizebox{0.95\textwidth}{!}{
    \begin{tabular}{lcccccccc}
        \toprule
        & \multicolumn{2}{c}{Goal Conditioning} & & \multicolumn{4}{c}{Metrics} \\
        \cmidrule(lr){2-3}
        \cmidrule(lr){5-8}
        \multicolumn{1}{c}{Variant}
        & Goal Token
        & $\alpha$
        &
        & ADE $(m)$ $\downarrow$
        & FDE@4s $(m)$ $\downarrow$
        & Goal Progress (GP) $\uparrow$
        & Ego Progress (EP) $\uparrow$ \\
        \midrule
        EP only
        & \checkmark
        & 0.0
        &
        & 0.251
        & 0.239
        & 0.979
        & 0.970 \\
        
        w/o Goal Token
        & $\times$
        & 0.5
        &
        & 4.364
        & 7.766
        & 0.718
        & 0.709 \\
        
        \midrule
        
        Go2-DrivoR
        & \checkmark
        & 0.5
        &
        & \textbf{0.202}
        & \textbf{0.207}
        & \textbf{0.988}
        & \textbf{0.976} \\
        
        Goal-only
        & \checkmark
        & 1.0
        &
        & 0.325
        & 0.463
        & 0.932
        & 0.901 \\
        \bottomrule
    \end{tabular}
    }
\end{table*}

\subsection{Open-loop Evaluation}
\subsubsection{Simulator}
We compare Go2-DrivoR with a constant-velocity extrapolation baseline and iPlanner~\cite{Yang-RSS-23}, and report the logged expert trajectory (GT Agent) as an oracle reference.
For iPlanner, we use the official pretrained checkpoint without fine-tuning, providing a front-facing pinhole depth image and the same local goal as Go2-DrivoR.
We convert its output path into a four-second trajectory by sampling at 0.5\,s intervals using the current ego speed, with headings determined by the path tangent.
After reaching the path endpoint, the position is held fixed.

We first evaluate Go2-DrivoR in an open-loop setting on TartanGround~\cite{patel2025tartanground} to assess trajectory prediction and scoring quality under both seen and unseen maps.
Following the standard benchmark protocol, we report no-fault collision (NOC), time-to-collision (TTC), drivable-area compliance (DAC), goal-conditioned ego progress (GEP), comfort, and an adapted PDMS-style~\cite{Dauner2024NEURIPS} aggregate score in Eq.~\eqref{eq:aggregate}.

Table~\ref{tab:ol_sim} shows that Go2-DrivoR achieves higher aggregate scores than both constant-velocity and zero-shot iPlanner on seen and unseen maps.
Compared with iPlanner, Go2-DrivoR achieves higher average NOC, TTC, DAC, and comfort scores on both splits.
While iPlanner achieves slightly higher GEP on unseen maps, Go2-DrivoR achieves a higher aggregate score across the evaluated criteria.

The qualitative example in Fig.~\ref{fig:ol_sim} shows the selected trajectory advancing toward the supplied local goal while remaining within the allowable region.
Although it differs from the logged expert path, it achieves an oracle-evaluated PDMS-style score of 0.933, illustrating that a high-scoring trajectory need not exactly reproduce the expert motion.


\subsubsection{Real-world}
We also evaluate zero-shot transfer on real-world logs collected with the Unitree Go2.
We evaluated without additional training on the real-world data, and the collected logs are used only for testing.
We report average displacement error (\texttt{ADE}), final displacement error at multiple horizons (\texttt{FDE@1s}, \texttt{FDE@2s}, and \texttt{FDE@4s}), and heading mean absolute error (\texttt{Heading MAE}).

As shown in Table~\ref{tab:ol_real}, Go2-DrivoR outperforms the constant-velocity baseline across all reported metrics and both scene categories.
As expected from conditioning on the four-second lookahead waypoint, the largest reduction occurs at FDE@4s; Go2-DrivoR also improves ADE, shorter-horizon FDEs, and Heading MAE.
The method produces trajectories that better match the ground-truth future motion and yields improved heading consistency, demonstrating effective zero-shot sim-to-real transfer.

The qualitative result in Fig.~\ref{fig:ol_real} further shows that the predicted trajectories are better aligned with the intended navigation direction.
In particular, in the intersection example, Go2-DrivoR selects a turning trajectory toward the goal, whereas the constant-velocity baseline continues along the initial heading, illustrating the effect of goal-conditioned trajectory selection.
Overall, these results demonstrate zero-shot transfer to goal-conditioned trajectory prediction on real-world logs under the evaluated open-loop protocol.

\subsection{Ablation Studies}
We investigate the contributions of explicit goal-token conditioning and progress blending.
All variants are trained exclusively on TartanGround and evaluated on the same real-world Unitree Go2 logs described in Section~\ref{subsubsec:data_real}, using the same open-loop evaluation protocol.
Results are averaged over all scenarios from both the pedestrian and road categories.
The goal-token ablation retains $\alpha=0.5$ in the GEP formulation of Eq.~\eqref{eq:gep}, whereas the progress ablations retain goal-token conditioning and vary $\alpha\in\{0,0.5,1\}$.

As shown in Table~\ref{tab:goal_ablation}, removing the goal token while retaining the blended progress objective leads to a substantial degradation in both trajectory prediction accuracy and goal-oriented metrics.
A possible explanation is that coarse directional commands leave local paths ambiguous in open urban spaces, whereas an explicit goal provides a precise spatial target.
However, it remains unclear why the ablated model exhibits substantially larger trajectory prediction errors than constant-velocity extrapolation despite receiving ego-motion inputs.
Further analysis is needed to determine whether the degradation arises primarily from candidate generation or trajectory scoring.

We further investigate the effect of the progress formulation by varying $\alpha$, which controls the balance between the original ego-progress target and goal-directed progress.
Both the EP-only and Goal-only variants perform worse than the blended formulation used in Go2-DrivoR.
In particular, relying solely on goal-directed progress does not provide the same trajectory accuracy or progress quality as combining it with the original ego-progress signal.
These results support combining the existing ego-progress signal with goal  progress in the evaluated setting.
\section{CONCLUSION}
\label{sec:conclusion}
We present Go2-DrivoR, a lightweight goal-conditioned adaptation of DrivoR for urban quadrupedal navigation.
By conditioning trajectory generation on a local-frame subgoal, redefining drivable-area compliance for sidewalk-oriented navigation, and reformulating ego progress as GEP, the method enables destination-based local planning without redesigning the core decoders.
Trained exclusively on TartanGround, Go2-DrivoR improves planning performance on unseen simulation maps and transfers zero-shot to real-world Go2 trajectory prediction.
Limitations include evaluation with expert-derived lookahead goals, reliance on open-loop real-world evaluation, and the lack of explicit dynamic-agent modeling.
Future work will focus on closed-loop social navigation adaptation, temporal scene modeling, and improved sim-to-real adaptation.



\setcounter{figure}{0}
\renewcommand{\thefigure}{A\arabic{figure}}

\section*{APPENDIX}
\label{sec:appendix}

\subsection{Preliminary: DrivoR}

DrivoR~\cite{kirby2026drivor} is a vision-centric end-to-end autonomous driving framework that maps multi-camera image inputs and ego states to a set of candidate trajectories and their scores.
It consists of a perception encoder, a trajectory decoder, and a disentangled scoring decoder.

\subsubsection{Visual Feature Extraction}
DrivoR uses a DINOv2-pretrained ViT-S backbone \cite{oquab2024dinov2}, based on the Vision Transformer (ViT) architecture~\cite{dosovitskiy2021an}, and fine-tunes it with LoRA~\cite{hu2022lora}.
Building on the register-token formulation of Darcet et al.~\cite{darcet2024vision}, DrivoR introduces camera-specific learnable registers that compress patch-level visual features into a compact scene representation.
Each camera is processed independently, and the resulting camera-aware register tokens are collected to form
\[
F \in \mathbb{R}^{N_c \times N_r \times D},
\]
where $N_c$ is the number of cameras, $N_r$ is the number of registers
per camera, and $D$ is the feature dimension.

\subsubsection{Trajectory Decoder}
Given the scene representation $F$, the trajectory decoder generates $K$ candidate trajectories using learnable trajectory queries $\{q_k\}_{k=1}^{K}$.
The original DrivoR formulation embeds ego-state information, including pose, velocity, acceleration, and driving command, and adds this embedding to the trajectory queries before they enter the decoder.
Each query then attends to $F$ through cross-attention and is decoded into a trajectory $\tau_k \in \mathbb{R}^{n_p \times 3}$ in the ego frame.
During training, the trajectory decoder is optimized using a winner-takes-all (WTA) regression objective, in which only the candidate closest to the expert trajectory is supervised.

\subsubsection{Scoring Decoder}
DrivoR employs a separate scoring decoder to evaluate the generated trajectory candidates.
Each decoded trajectory $\tau_k$ is detached from the trajectory-generation branch, re-embedded into a scoring query using an MLP, and processed by a transformer decoder that cross-attends to the same scene representation $F$.
This separation prevents scoring gradients from propagating into the trajectory decoder while still allowing the perception encoder to learn features useful for both generation and scoring.

The scoring decoder predicts six component-wise scores derived from the Predictive Driver Model Score (PDMS)~\cite{Dauner2024NEURIPS}: no-fault collision (NOC), drivable-area compliance (DAC), driving-direction compliance (DDC), ego progress (EP), time-to-collision (TTC), and comfort.
The individual components are supervised using binary cross-entropy against oracle scores.
Although DDC is predicted and supervised, its aggregation weight is zero in the NAVSIM-v1 PDMS formulation used by DrivoR.
At inference, the predicted sub-scores are combined to rank the candidate trajectories, and the highest-scoring candidate is selected.

\subsection{Related Work}

\subsubsection{Robot Navigation}
Classical robot navigation systems typically decompose perception, mapping, planning, and control, with representative local-planning approaches including artificial potential fields~\cite{1087247} and the dynamic window approach~\cite{580977}.
Learning-based approaches have increasingly replaced parts of this pipeline with policies or planners that directly predict local paths from perceptual observations and navigation goals.

iPlanner~\cite{Yang-RSS-23} predicts goal-directed local paths from depth observations using imperative learning, while ViPlanner~\cite{roth2024viplanner} extends this paradigm by incorporating geometric and semantic information for terrain-aware local navigation.
Beyond local planning, ViNT~\cite{shah2023vint} learns a general-purpose visual navigation policy from diverse cross-robot navigation data, and NoMaD~\cite{sridhar2023nomad} extends goal-conditioned visual navigation with a diffusion policy and goal masking to unify goal-reaching and exploration.

Language-conditioned embodied agents have also incorporated semantic instructions and pretrained models for navigation and task completion~\cite{suglia2021embodied,shah2022lmnav}.
More recently, navigation foundation models such as NavFoM~\cite{zhang2025embodied} aim to transfer navigation knowledge across tasks and robot embodiments.
In contrast, our work focuses on short-horizon quadrupedal local planning in which multiple goal-conditioned trajectories are explicitly generated and ranked using disentangled criteria for safety, drivability, and progress toward a supplied local goal.

\subsubsection{End-to-End Autonomous Driving}
End-to-end autonomous driving has increasingly adopted architectures that map multi-modal sensor observations directly to planned trajectories.
TransFuser~\cite{Chitta2023PAMI} integrates visual and LiDAR features for end-to-end perception and planning, while diffusion-based planners model multi-modal future trajectory distributions and enable flexible trajectory generation \cite{yang2024diffusiones,diffusiondrive,zou2025diffusiondrivev2,zheng2025diffusion}.

SparseDrive~\cite{sun2025sparsedrive} demonstrates efficient end-to-end planning using sparse scene representations, whereas DrivoR~\cite{kirby2026drivor} uses compact camera-aware register tokens together with separate trajectory-generation and trajectory-scoring decoders.

These methods are primarily designed for vehicle-centric navigation.
In particular, the PDMS-based scoring formulation adopted by DrivoR uses vehicle-oriented drivable-area and ego-progress criteria that do not explicitly measure advancement toward an arbitrary local destination coordinate.
Our work adapts this candidate-generation and scoring paradigm to quadrupedal navigation by conditioning trajectory proposals on a local-frame goal and incorporating goal-directed progress into trajectory evaluation.

\subsubsection{Evaluation Platforms and Datasets}
Urban navigation and autonomous driving policies are commonly evaluated using either non-reactive logged-data protocols or interactive closed-loop simulation.
In non-reactive evaluation, the policy output does not affect subsequent sensor observations or environment states.
NAVSIM~\cite{Dauner2024NEURIPS}, for example, uses a non-reactive simulation over logged driving scenes to compute safety-, comfort-, and progress-oriented metrics for predicted trajectories.

For ground-robot research, TartanGround~\cite{patel2025tartanground} provides a large-scale multimodal dataset collected in diverse photorealistic simulation environments, including multi-camera images, poses, inertial measurements, and semantic information.
In this work, we process TartanGround into short-horizon local planning samples and use the resulting data for training and non-reactive open-loop evaluation.

By contrast, closed-loop simulators feed policy outputs back into the environment so that subsequent observations and interactions depend on the agent's behavior.
CARLA~\cite{dosovitskiy2017carla} provides interactive urban simulation for autonomous driving, while MetaUrban~\cite{wu2025metaurban} targets urban micromobility and supports point- and social-navigation tasks in procedurally generated interactive environments with heterogeneous mobile agents.
These platforms provide complementary tools for interactive evaluation, whereas the present work focuses on non-reactive trajectory evaluation and leaves closed-loop quadrupedal navigation for future study.

\section*{ACKNOWLEDGMENT}
This work was supported by the Korea Institute of Science and Technology (KIST) Institutional Program (Project No. 26E0061).



\bibliographystyle{IEEEtran}
\bibliography{references}

\end{document}